\documentclass{article}

\usepackage{arxiv}

\usepackage[utf8]{inputenc} % allow utf-8 input
\usepackage[T1]{fontenc}    % use 8-bit T1 fonts
\usepackage{hyperref}       % hyperlinks
\usepackage{url}            % simple URL typesetting
\usepackage{booktabs}       % professional-quality tables
\usepackage{amsfonts}       % blackboard math symbols
\usepackage{nicefrac}       % compact symbols for 1/2, etc.
\usepackage{microtype}      % microtypography
\usepackage{lipsum}		% Can be removed after putting your text content
\usepackage{graphicx}
\usepackage{natbib}
\usepackage{doi}

\usepackage{subcaption}
\usepackage{amsmath}
\usepackage{multirow}
\usepackage{placeins}

\title{Decoding the Past: Explainable Machine Learning Models for Dating Historical Texts}

\author{%
Paulo J. N. Pinto$^{1,2,3,*}$, Armando J. Pinho$^{1,2,3}$, and Diogo Pratas$^{1,2,3,4,*}$ \\[0.5em]
{\small\begin{minipage}{\linewidth}\begin{center}
\begin{tabular}{ccc}
$^1$IEETA - Institute of Electronics and Informatics Engineering of Aveiro, and\\
$^2$LASI - Intelligent Systems Associate Laboratory, and \\
$^3$DETI - Department of Electronics, Telecommunications and Informatics,\\ University of Aveiro, 3810-193 Aveiro, Portugal \\
$^4$DoV - Department of Virology, University of Helsinki, 00014 Helsinki, Finland\\
$^*$Corresponding authors: paulojnpinto02@ua.pt and pratas@ua.pt\\
\end{tabular}
\end{center}\end{minipage}}
}

\date{}

\renewcommand{\shorttitle}{Explainable Machine Learning Models for Dating Historical Texts}

\hypersetup{
pdftitle={Explainable Machine Learning Models for Dating Historical Texts},
pdfsubject={Dating Historical Texts},
pdfauthor={Paulo J. N. Pinto, Armando J. Pinho, Diogo Pratas},
pdfkeywords={Machine Learning, Supervised Models, Text Dating, Compression-based features, Explainable AI},
}

\begin{document}
\maketitle

\begin{abstract}
Accurately dating historical texts is essential for organizing and interpreting cultural heritage collections. This article addresses temporal text classification using interpretable, feature-engineered tree-based machine learning models. We integrate five feature categories—compression-based, lexical structure, readability, neologism detection, and distance features—to predict the temporal origin of English texts spanning five centuries. Comparative analysis shows that these feature domains provide complementary temporal signals, with combined models outperforming any individual feature set.
On a large-scale corpus, we achieve 76.7\% accuracy for century-scale prediction and 26.1\% for decade-scale classification, substantially above random baselines (20\% and 2.3\%). Under relaxed temporal precision, performance increases to 96.0\% top-2 accuracy for centuries and 85.8\% top-10 accuracy for decades. The final model exhibits strong ranking capabilities with AUCROC up to 94.8\% and AUPRC up to 83.3\%, and maintains controlled errors with mean absolute deviations of 27 years and 30 years, respectively. For authentication-style tasks, binary models around key thresholds (e.g., 1850–1900) reach 85–98\% accuracy.
Feature importance analysis identifies distance features and lexical structure as most informative, with compression-based features providing complementary signals. SHAP explainability reveals systematic linguistic evolution patterns, with the 19th century emerging as a pivot point across feature domains. Cross-dataset evaluation on Project Gutenberg highlights domain adaptation challenges, with accuracy dropping by 26.4 percentage points, yet the computational efficiency and interpretability of tree-based models still offer a scalable, explainable alternative to neural architectures.
\end{abstract}

% keywords can be removed

\section{Introduction}
\label{sec:intro}

Historical texts serve as windows into past civilizations, revealing the diplomatic relations and societal changes that shaped our world. When we can accurately date these documents, we unlock the chronological frameworks needed to interpret history ~\citep{alkhalifa2023building,grabovoy2021automatic}. The Treaty of Tordesillas (1494) exemplifies this importance, dividing the New World between Spain and Portugal. Its precise dating helped resolve territorial disputes that lasted centuries. The original documents, now preserved at the General Archive of the Indies in Spain and Torre do Tombo National Archive in Portugal, earned UNESCO's Memory of the World recognition for their lasting historical significance \citep{unesco2007tordesillas}.

Ancient diplomatic correspondence tells similar stories. The Amarna Letters (c. 1350-1330 BCE) represent humanity's earliest international diplomatic system, 382 cuneiform tablets documenting exchanges between Egyptian pharaohs and neighboring rulers \citep{moran1992amarna, rainey2014amarna}. Through dating based on internal evidence spanning from Amenhotep III's final decade to Tutankhamun's early reign, historians reconstructed ancient geopolitical relationships and diplomatic practices. The Merneptah Stele (c. 1213 BCE) offers another example: containing the earliest textual reference to Israel in Egyptian records, it demonstrates how dated inscriptions anchor our understanding of ancient Near Eastern history \citep{kitchen2003pharaoh}.

Yet dating matters beyond diplomatic history because it shapes our understanding of literary evolution, legal precedents, and cultural development. Modern computational approaches achieve up to 79\% accuracy when dating ancient Greek inscriptions, though fragmentary texts still pose challenges \citep{sommerschield2023machine}. Archaeological discoveries continue revealing dated manuscripts that reshape our view of ancient civilizations. Meanwhile, misattributed or incorrectly dated texts have caused historical misinterpretations, underscoring why we need reliable, automated methods when traditional paleographic approaches fall short \citep{kestemont2017collaborative}.

Automatically predicting when historical texts were written spans computational linguistics, pattern recognition, and digital humanities, a challenge growing more urgent as digital repositories swell with millions of historical documents. Accurately determining when a text was written proves essential for historical research, literary analysis, plagiarism detection, and tracking temporal trends.

Traditional approaches to temporal text classification have depended on linguistic expertise and manual feature engineering. Researchers must possess deep domain knowledge to spot relevant temporal markers, vocabulary shifts, syntactic changes, and evolving stylistic conventions \citep{ koppel2009computational}. \cite{luyckx2008authorship} demonstrated how manually crafted stylometric features excel in historical text analysis. Though interpretable, these methods struggle with scalability and require extensive preprocessing for each new corpus or time period \citep{qian2022survey}.

Large-scale collections from Project Gutenberg and Open Library/Internet Archive have opened new possibilities for automated temporal classification. These repositories house document collections spanning primarily the 17th to 21st centuries, rich with temporal signals across diverse literary genres and writing styles. To leverage these collections effectively, we need computational approaches that capture temporal variation's complexity while remaining both accurate and interpretable.

Text classification has recently advanced through two contrasting paths. Neural architectures have reached new performance heights, with \cite{devlin2018bert} setting standards on GLUE and SQuAD datasets, \cite{liu2019roberta} achieving 94.54\% accuracy on classification tasks, and \cite{raffel2020exploring} demonstrating T5's superiority with an 88.9 score on SuperGLUE \citep{rogers2020primer}. Yet simple, theoretically-grounded methods using tree-based models and compression algorithms have experienced a renaissance, often matching or surpassing complex neural systems while demanding far fewer computational resources.

Our approach addresses this challenge by combining compression-based features with linguistic characteristics, using interpretable tree-based models to predict text age. We leverage normalized compression metrics from Markov chain models to capture how text predictability patterns shift over time, while extracting lexical, structural, and readability features that track vocabulary evolution, syntactic complexity, and changing stylistic conventions across historical periods.

This work makes three contributions. First, we develop a method for automatically dating historical texts by integrating compression-based features with linguistic characteristics through interpretable tree-based models, achieving good accuracy while requiring far fewer computational resources than neural approaches. Second, our feature importance analysis reveals how specific linguistic patterns evolve across decades and centuries. Third, we establish an evaluation framework using large-scale historical corpora that addresses the temporal distribution challenges inherent in historical text analysis.

Our evaluation uses English texts spanning five centuries, with filtering to ensure data quality and adherence to FAIR principles (Findable, Accessible, Interoperable, Reusable) established by the European Union's research data management guidelines \citep{wilkinson2016fair}. 

The paper proceeds as follows: Section~\ref{sec:related_work} reviews related work in temporal text classification and compression-based text analysis; Section~\ref{sec:methodology} details our feature extraction and tree-based modeling approach; Section~\ref{sec:feature_benchmark} presents experimental evaluation with feature importance analysis; Section~\ref{sec:results} presents results and implications, while Section~\ref{sec:discussion} discusses the results; Section~\ref{sec:conclusion} concludes with future research directions.

%%%%%%%%%%%%%%%%%%%%%%%%%%%%%%%%%%%%%%%%%%%%%%%%%%%%%%%%%%

\section{Related Work}
\label{sec:related_work}

\subsection{Temporal Text Classification}

Temporal text classification has evolved from rule-based linguistic approaches toward machine learning systems. Early researchers concentrated on identifying explicit temporal markers and stylistic features that correlate with specific historical periods~\citep{niculae2014temporal,dalli2006automatic}. Though these methods offered clear interpretability, they required extensive manual feature engineering and domain expertise.

Recent advances have been dominated by neural approaches. Time-aware language models and transformer-based architectures now achieve state-of-the-art performance of 85-92\% accuracy on benchmark datasets~\citep{meng2023temporal, kong2024time}. These models integrate temporal information directly into their embedding processes, addressing the challenge of semantic drift as words evolve meaning over time.

These approaches, however, come with trade-offs. Neural models struggle with computational scalability and suffer from temporal drift, showing 10-45\% accuracy degradation when applied across different time periods~\cite{alkhalifa2023building}. Pre-trained models experience temporal effects that limit their applicability to historical text analysis~\citep{lazaridou2021temporal}. These approaches sacrifice interpretability, making it difficult to understand which linguistic features actually drive temporal discrimination or how language evolution mechanisms operate.

\subsection{Compression-Based Text Analysis}

Compression-based text analysis has gained renewed interest, drawing on principles from information theory and Kolmogorov complexity \citep{li1997kolmogorov}. The theoretical foundation rests on a key measure: normalized relative compression (NRC) \citep{ziv2002measure,pinho2016authorship}. Both measures leverage the principle that compression algorithms can approximate optimal compressors, creating universal similarity measures that work across domains without requiring task-specific training. The approach is based on the normalized information distance framework \citep{li2004similarity}, which establishes a theoretical basis to measure the similarity between arbitrary objects using compression algorithms as approximations of the Kolmogorov complexity. \cite{pinho2016authorship} provided evidence of NRC's effectiveness in authorship attribution, achieving very high classification accuracy through relative compression techniques that measure how much of a target text can be reconstructed using information from a reference text \cite{pinho2018application}.

\cite{jiang2023low} showed that simple compression algorithms combined with k-nearest neighbor classification can compete directly with transformer models while requiring orders of magnitude fewer computational resources. Their parameter-free method delivered competitive performance across multiple languages and domains, challenging the assumption that effective text classification requires complex neural architectures. Recent work has continued to validate this paradigm shift. AIDetx \citep{Almeida2024}, for example, applies compression-based classification to AI-generated text detection and achieves F1 scores exceeding 97\% while demanding substantially less computational power than traditional deep learning approaches.

Theoretical developments have strengthened this paradigm shift by establishing clear equivalence between language modeling and compression~\citep{deletang2023language}. Although \cite{shannon1948mathematical} established the theoretical connection between optimal prediction and optimal compression in information theory work, \cite{deletang2023language} provided empirical validation using modern large language models. Early applications of this principle to text classification include \cite{teahan2000text} and \cite{frank2000text}, who demonstrated compression-based approaches for text categorization. \cite{deletang2023language} showed that large language models function as powerful general-purpose compressors, with models like Chinchilla 70B achieving compression rates that exceed traditional algorithms while simultaneously serving as predictive models.

Compression-based approaches offer unique advantages for temporal applications: they naturally capture sequential dependencies through text processing, handle arbitrary-length patterns, and remain language independent with minimal preprocessing requirements. Recent work has shown compression-based methods achieving 73-75\% F1-scores on standard benchmarks while requiring no training parameters.

\subsection{Tree-Based Models in Text Classification}

Large-scale comparative studies have revealed competitive performance from tree-based ensemble methods in tabular data tasks. \cite{borisov2022deep} conducted an evaluation across 45 tabular datasets spanning classification, regression, and ranking problems, demonstrating that tree-based models frequently outperform deep learning approaches. This advantage becomes pronounced on datasets with mixed feature types and irregular distributions, characteristics directly relevant to temporal text data where linguistic features combine with metadata and temporal indicators.

\cite{grinsztajn2022tree} provided theoretical insights explaining why tree-based models excel on tabular data. They showed that XGBoost and CatBoost achieve better performance compared to deep learning methods including MLPs, ResNet, and FT\_Transformer across 45 diverse datasets, primarily through their effective handling of feature interactions and irregular patterns. Their analysis revealed that gradient boosting methods prove particularly well-suited for datasets where deep learning's inductive biases misalign with the underlying data structure, especially when dealing with uninformative features and non-rotationally invariant data.

\cite{mosbach2024xgboost} demonstrated a case where XGBoost outperformed GPT-4 on text classification while requiring $3{,}700\times$ less memory. This finding highlights the practical advantages of tree-based approaches: computational efficiency, interpretability through feature importance rankings, and robustness to distribution shifts.

Tree-based models offer specific advantages for temporal text classification: built-in feature importance for understanding temporal patterns, robust handling of mixed data types (text features and temporal metadata), linear scaling with data size, and superior deployment characteristics for production systems processing continuous text streams.

\subsection{Model Interpretability and Feature Analysis}

Understanding which features drive temporal discrimination requires analytical approaches that go beyond classification accuracy. Tree-based models offer built-in feature importance mechanisms that rank features according to their contribution to predictive performance across the entire model ensemble. Permutation importance analysis complements this approach by quantifying feature relevance through measuring performance degradation when individual features are randomly shuffled, providing robust validation of feature rankings. These interpretability methods prove valuable for temporal text classification because they identify which linguistic features contribute to predictions while revealing how individual characteristics combine to drive specific temporal classifications. Such interpretability offers insights into temporal discrimination mechanisms and validates that model decisions align with linguistic expectations about historical language evolution.

\subsection{Historical Text Corpora}

The availability of large-scale historical text collections has been important for temporal text classification research. Project Gutenberg remains the most used corpus, with the Standardized Project Gutenberg Corpus (SPGC)~\cite{strobl2020spgc} containing over 50,000 books with $3 \times 10^9$ word-tokens across more than 20 languages, spanning multiple centuries. The SPGC encompasses diverse text types including novels, poetry, short stories, dramas, biographies, essays, speeches, letters, and reference works, with systematic categorization through bookshelf metadata covering genres from philosophy to children's literature. The Open Library/Internet Archive collections complement this with broader historical materials including legal documents, judicial records, newspapers, periodicals, government documents, and manuscripts, providing richer temporal and topical diversity. However, researchers consistently note challenges including manual selection bias, non-uniform topical composition over time, and concentration of pre-1930 content due to copyright restrictions.

~\cite{gerlach2013stochastic} provided foundational analysis of vocabulary growth patterns in natural languages using Project Gutenberg data, establishing statistical models that inform our understanding of lexical evolution over time. Their work demonstrates that historical text collections contain quantifiable temporal signals that can be leveraged for classification tasks.

Open Library usage in temporal text classification research appears limited based on recent literature, with most researchers preferring institutional digital libraries or specialized historical corpora with better temporal metadata and preprocessing standards. This presents an opportunity for research that effectively leverages Open Library's extensive collection alongside established corpora like Project Gutenberg.

\subsection{Research Gaps and Positioning}

Despite these advances, current approaches face limitations that create research opportunities. Neural methods struggle with computational scalability and temporal drift, while traditional linguistic approaches lack automation and scalability. Compression-based methods show promise but haven't been systematically combined with complementary linguistic features for temporal classification tasks.

Our approach addresses these gaps by combining the theoretical foundations of compression-based analysis with the practical advantages of tree-based models and the interpretability of explicit linguistic features. This integration offers a solution that balances accuracy, efficiency, and interpretability while leveraging large-scale historical corpora for robust temporal text classification.

%%%%%%%%%%%%%%%%%%%%%%%%%%%%%%%%%%%%%%%%%%%%%%%%%%%%%%%%%%

\section{Methodology}
\label{sec:methodology}

The methodology comprises four components: dataset construction and preprocessing, feature extraction, tree-based modeling with explainability analysis, and evaluation frameworks for multiclass and binary classification. Within the feature extraction component, we consider five feature categories—compression-based, lexical structure, readability, neologism detection, and function-word distance features—that capture complementary statistical and linguistic signals. 

Multiclass classification involves predicting the specific temporal period of a text, such as the exact decade (e.g., 1850s, 1860s, 1870s) or century (e.g., 17th, 18th, 19th, 20th, 21st) when it was written. Binary classification addresses temporal boundary detection, determining whether a text was written before or after a specific temporal threshold (e.g., older than 1900, newer than 1850). Multiclass methods enable fine-grained temporal localization, while binary methods identify broad historical transitions and temporal boundaries.

\subsection{Dataset Construction and Preprocessing}

Dataset construction requires attention to temporal balance, source diversity, and preprocessing consistency. Historical text digitization presents challenges, including OCR artifacts, encoding inconsistencies, and imbalances in temporal distribution, which can affect classification performance. We acquired historical texts from multiple repositories, implemented preprocessing pipelines, and ensured data quality suitable for feature extraction and machine learning analysis. 

\subsubsection{Multisource Data Acquisition Strategy}

Creating a balanced temporal corpus requires acquiring texts from two sources: Project Gutenberg and Open Library/Internet Archive. This dual-source approach mitigates limitations of individual repositories while maximizing temporal coverage and diversity.

For the Project Gutenberg collection, we used automated downloading with \texttt{wget}, respecting platform usage policies by including appropriate request delays.

For the Open Library collection, a custom Python script queries the Open Library search API, filters for publicly accessible English-language books with confirmed publication dates, and downloads OCR-derived text files from Internet Archive.

All texts were collected with the intention of obtaining the original publication dates to ensure accurate temporal classification. However, some texts, particularly within the Project Gutenberg collection, may contain dates corresponding to new editions or composition dates rather than original first publication dates due to the complex bibliographic history of historical works or lack of information about the original publication date in the sources. To maintain data quality, texts with ambiguous or uncertain publication dates were systematically discarded during the initial data curation phase.

The resulting corpus encompasses over 200,000 texts spanning from the 15th to 21st centuries, with temporal metadata enabling both century-scale and decade-scale classification tasks. To address class imbalance in historical text preservation, we used stratified sampling with size limits: 500 MB per decade for Open Library and 50 MB per decade for Project Gutenberg, prioritizing texts exceeding 1 MB for sufficient content. This sampling strategy yielded 17,247 texts from Open Library/Internet Archive and 1,737 texts from Project Gutenberg, for a final dataset of 18,984 texts.

\subsubsection{Text Preprocessing Pipeline}

The preprocessing pipeline implements standardized text cleaning while preserving linguistic characteristics essential for temporal classification. This includes removing Project Gutenberg headers and footers, converting to UTF-8 encoding, and normalizing whitespace characters while preserving linguistic content and punctuation.

Dataset organization uses hierarchical chronological structures enabling both century-level and decade-level analysis. Project Gutenberg and Open Library datasets are maintained as separate corpora throughout the study to enable independent evaluation and cross-dataset validation. The Open Library texts are partitioned into training (65\%), validation (25\%), and test (10\%) sets using stratified sampling to maintain temporal distribution balance, while the Project Gutenberg texts serve as an external validation corpus for assessing generalization capabilities across different historical text collections.

\subsection{Feature Extraction Framework}

Feature extraction generates diverse quantitative descriptors organized into five categories capturing different aspects of temporal linguistic variation. This approach enables tree-based models to identify characteristic combinations that distinguish between temporal periods.

\subsubsection{Compression-Based Features}

Compression-based features exploit the principle that texts with different predictability exhibit varying compressibility. These features capture statistical and structural properties that correlate with historical writing conventions and linguistic evolution.

Shannon Entropy provides the fundamental statistical uncertainty baseline for character distribution analysis. Shannon entropy is given by
\begin{equation}
H_{\text{Shannon}} = -\sum_{c \in A} \frac{n_c}{N} \log_2 \left(\frac{n_c}{N}\right),
\end{equation}
where $n_c$ represents the count of character $c$, $N$ is the total number of characters, and the summation is over all characters $c$ in the alphabet $A$. Higher values indicate more uniform character distributions and lower predictability~\citep{shannon1948mathematical}.

The Entropy Ratio compares entropy estimated from Markov, prediction by partial matching or finite-context models with Shannon entropy as
\begin{equation}
\text{Entropy Ratio} = \frac{H_{\text{Markov}}}{H_{\text{Shannon}}}.
\end{equation}
This ratio quantifies how well sequential dependencies captured by Markov models explain character distribution entropy, revealing temporal patterns in linguistic structure.

Normalized Relative Compression (NRC) serves as our primary relative compression-based metric. In its general form, NRC is defined as
\begin{equation}
\text{NRC}(x||y) = \frac{C(x||y)}{|x| \cdot \log_2 |A|},
\end{equation}
where $C(x||y)$ represents the compression of text $x$ using exclusively the compression model built from reference text $y$, $|x|$ denotes character count, and $\log_2 |A|$ represents per-character entropy of a uniformly random string over the alphabet $A$. This measures how much of the target text, $x$, can be constructed using information from a reference text, $y$.

Additional compression features include Compression Ratio calculated using Variable Order Markov Models (VOMM) implementing Prediction by Partial Matching (PPM) algorithms. These algorithms dynamically adjust context windows to capture patterns ranging from local character dependencies to broader stylistic signatures, revealing temporal signatures in text predictability through adaptive compression analysis.

\subsubsection{Lexical and Structural Features}

Lexical and structural features quantify stylistic conventions, syntactic complexity, and formatting characteristics that vary across historical periods. These features capture observable textual properties reflecting stylistic choices and linguistic trends.

Key lexical features include Average Word Length, Lexical Richness (type-token ratio measuring vocabulary diversity)~\citep{koppel2009computational}, Average Sentence Length, and Syllables per Word.

Structural features include Punctuation Density, Uppercase Ratio, and Digit Ratio~\citep{luyckx2008authorship}. These characteristics capture formatting conventions and composition patterns that may vary temporally.

\subsubsection{Neologism Detection Features}

The neologism detection system tracks vocabulary evolution across eight historical periods aligned with technological, social, and cultural developments. Each era generated distinctive linguistic signatures: scientific terminology in the Early Modern Period (1600-1749), industrial vocabulary in the 18th-19th centuries, and digital lexicons in contemporary communication. The vocabulary lists for each historical period were curated by the authors of this paper, with all selected words confirmed by them as created within their respective temporal boundaries to ensure chronological accuracy and prevent anachronistic vocabulary assignments. A complete table of neologisms for all eight historical periods, along with additional information, is available in Supplementary Section~1.

The system generates eleven features through analysis of vocabulary presence within texts, including period-specific boolean indicators, and a continuous vocabulary modernity score measuring the proportion of modern-period vocabularies present in texts. Implementation prevents data leakage by detecting vocabulary presence without judging historical appropriateness, avoiding circular reasoning while preserving discriminative power.

\subsubsection{Language-Dependent Readability Features}

Readability features assess cognitive accessibility and complexity using linguistic metrics that reflect historical trends in writing style, educational standards, and audience targeting. The Flesch Reading Ease score provides difficulty measurement for English texts based on word and sentence length characteristics~\citep{flesch1948new}, while Stopword Ratio measures grammatical function word proportion, indicating syntactic density~\citep{koppel2009computational}.

\subsubsection{Distance Features}

Distance (word) features examine grammatical evolution through positional analysis of 17 grammatical elements representing core syntactic functions. The analysis tracks positioning patterns for prepositions (at, by, for, in, of, on, to, with), articles (the, a, an), auxiliary verbs (is, was), coordinating elements (and), and complementizers/pronouns (as, that, it). These function words undergo positional transformations that yield temporal signatures reflecting changes in syntactic complexity, grammatical preferences, and stylistic conventions across historical periods. For each target function word, the system calculates mean distances between successive occurrences, capturing evolutionary patterns in preposition usage, agentive expression development, auxiliary verb positioning, coordination strategies, and subordination complexity that reflect syntactic development across centuries.

\subsection{Tree-Based Modeling Strategy}

The modeling approach uses gradient boosting algorithms that provide temporal predictions and interpretable insights into linguistic characteristics driving temporal discrimination. We employ CatBoost~\citep{prokhorenkova2018catboost} and XGBoost~\citep{chen2016xgboost} as primary algorithms, selected for their strengths in handling mixed feature types and providing feature importance analysis.

\subsubsection{Model Architecture and Training}

The tree-based architecture processes texts through feature representations, enabling evaluation of which linguistic characteristics distinguish between temporal periods. 

Training encompasses three formulations: century-scale classification (5 classes: 17th-21st centuries), decade-scale classification (43 classes: 1600s-2020s), and binary classification (also called authentication) using temporal thresholds. The binary approach transforms multiclass problems into threshold-based decisions, enabling evaluation of temporal discrimination across different historical boundaries.

\subsubsection{Feature Importance and Explainability Analysis}
The methodology incorporates feature importance analysis through tree-based importance mechanisms, permutation importance evaluation, and SHAP (SHapley Additive exPlanations) analysis. Tree-based models provide built-in feature ranking based on information gain and split frequency across ensemble trees, while permutation importance quantifies feature relevance by measuring performance degradation when individual features are randomly shuffled. SHAP analysis provides local and global explanations by computing the contribution of each feature to individual predictions and overall model behavior.
This triple-approach analysis provides insights into which linguistic characteristics drive temporal classification while validating feature importance rankings across different interpretability perspectives, enabling both aggregate understanding of temporal patterns and detailed examination of individual text classifications.

\subsection{Binary Classification Framework}

The methodology incorporates binary classification as an approach to multiclass temporal prediction. Binary classification addresses temporal boundary detection by determining whether texts were written before or after specific temporal thresholds. This approach provides different perspectives compared to fine-grained multiclass prediction, enabling investigation of broad historical transitions and temporal boundaries.

The binary classification framework uses identical feature representations and preprocessing pipelines as multiclass models. For each threshold year $t$, texts are labeled as "old" (written before year $t$) or "new" (written after year $t$). All algorithms use the same hyperparameter configurations as multiclass models to ensure fair comparison. Threshold values vary in decade or century increments to cover the full temporal range, enabling analysis of temporal discrimination across different historical periods.

Binary classification evaluation calculates classification metrics for each threshold configuration, revealing how temporal boundaries affect discrimination performance. This approach provides insights into temporal resolution limitations and identifies periods where linguistic evolution creates strong or weak classification signals.

\subsection{Evaluation Framework}

Model evaluation uses metrics assessing different aspects of temporal classification accuracy. For multiclass formulations, we use Accuracy, weighted F1-score~\citep{powers2011evaluation}, Area Under ROC Curve (AUCROC)~\citep{fawcett2006introduction}, Area Under Precision-Recall Curve (AUPRC)~\citep{davis2006relationship}, Mean Absolute Error (MAE), Root Mean Square Error (RMSE)~\citep{willmott2005advantages}, and Top-k Accuracy. Top-k metrics prove relevant for temporal classification, as texts from adjacent periods may share characteristics due to gradual linguistic evolution.

Binary classification evaluation uses the same metrics across threshold variations, enabling identification of optimal temporal boundaries for discrimination tasks.

%\subsubsection{Alphabet Normalization Impact Assessment}

%To quantify the benefits of character set reduction, we compared the full alphabet processing and lowercase-only variants. This analysis evaluated whether capitalization information contributes sufficiently to temporal discrimination to justify increased computational requirements during feature extraction.

%The alphabet comparison provides insights for deployment scenarios where computational efficiency is prioritized, while maintaining assessment of information loss through simplified character representations.

\subsection{Software Implementation}

The complete implementation of the temporal text classification methodology has been developed and made publicly available to ensure reproducibility and facilitate future research. The software package includes all feature extraction algorithms, tree-based model implementations, evaluation frameworks, and data preprocessing pipelines described in this methodology.

The source code, documentation, and scripts are available in a public GitHub repository, under the GPLv3 license, at \url{https://github.com/TextDate/Feature-Benchmark}. 

\subsection{Computational Resources}

We benchmarked the feature extraction pipeline in terms of runtime and memory usage across all dataset splits (training, validation, test, and Gutenberg). End-to-end feature extraction completed in approximately 4.7 hours, of which about 3.6 hours correspond to actual file processing and 1.1 hours to overhead tasks such as reference model fitting, initialization, and data cleaning.

Resource usage followed two regimes. During training and validation, we used a highly parallel configuration with 30 worker processes, which maximized throughput but led to higher aggregate memory consumption. In this regime, the training split exhibited the highest sustained usage, with total memory hovering around 150--165~GB due to the combination of many concurrent workers and large batches of files. For the test and Gutenberg splits, we adopted a reduced configuration with 4 worker processes to better reflect inference-phase deployment, resulting in lower coordination overhead and more concentrated memory usage per worker but a similar overall peak envelope.

Peak memory usage during distributed feature extraction reached approximately 165~GB, while reference model fitting for second-order Markov models produced the largest transient spikes at roughly 176~GB. These spikes occur only once per configuration, before steady-state processing. In practice, allocating 190--200~GB of RAM is sufficient to run the full pipeline with a comfortable safety margin. This characterization indicates that, while the feature extraction framework is computationally intensive at scale, it remains feasible on modern multi-core servers and can be downscaled (e.g., fewer workers, smaller batches) for more constrained environments at the cost of longer runtimes. Moreover, as discussed in the next section, some features with lower contribution to the final model can be disabled, substantially reducing both memory consumption and processing time.

\section{Feature Benchmark Analysis}
\label{sec:feature_benchmark}

This analysis examines five feature domains—compression, lexical structure, distance, neologism, and readability—to determine their individual contributions and best combinations for temporal classification at decade and century levels.

\subsection{Compression Features Domain Analysis}

Language evolution affects more than vocabulary and grammar; changes appear in text compressibility and entropy patterns. This principle enables temporal classification by identifying information-theoretic signatures unique to different historical periods. Compression features include Normalized Relative Compression (NRC) using first and second-order Markov models, Shannon entropy calculations, entropy ratios for each model order, and compression ratios for both orders.

\begin{figure}[!h]
\centering
\includegraphics[width=1\textwidth]{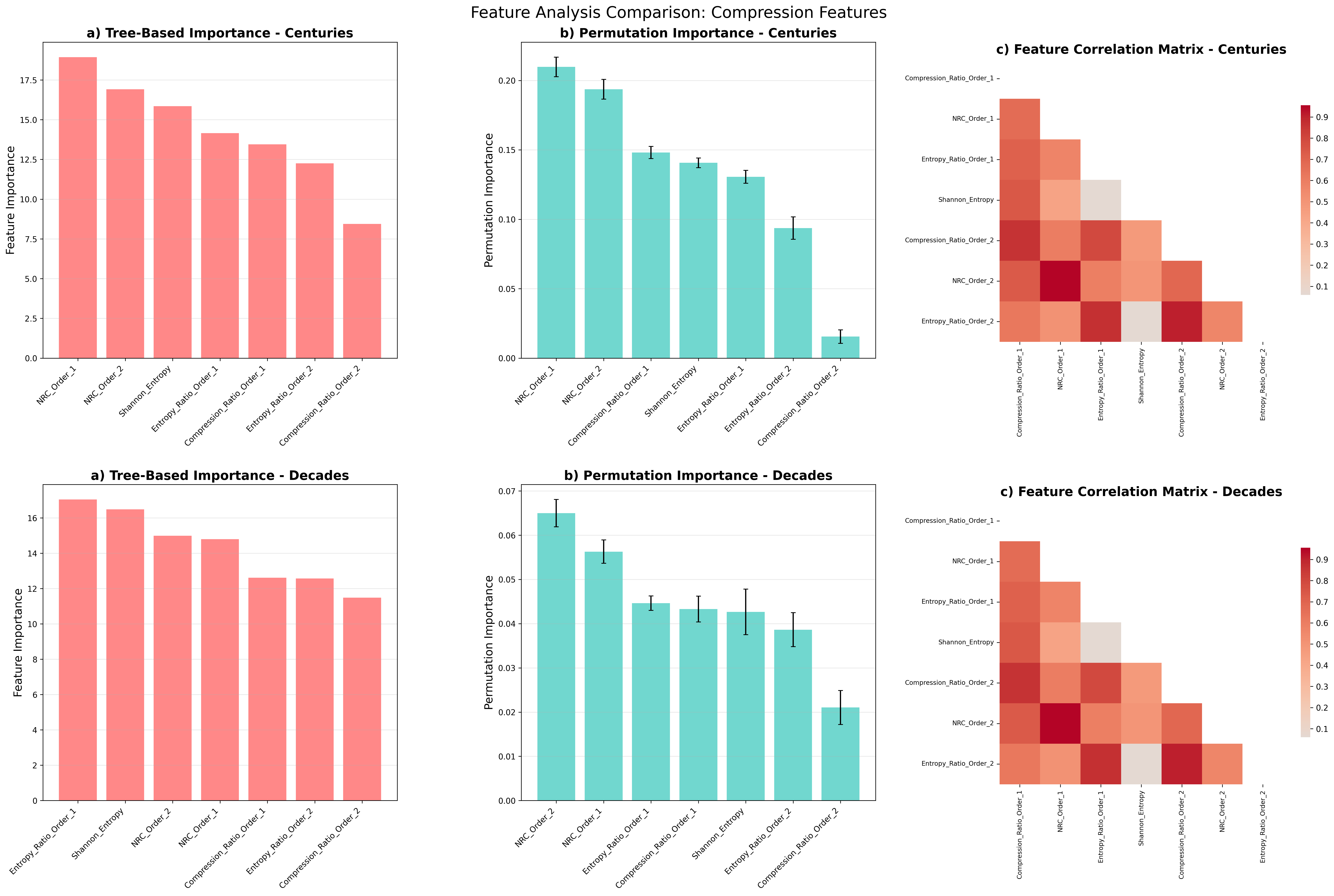}
\caption{Compression features classification: a) Tree-based importance, b) Permutation importance, c) Feature correlations.}
\label{fig:compression_combined_analysis}
\end{figure}

Figure~\ref{fig:compression_combined_analysis} demonstrates temporal-scale dependencies in feature effectiveness through comparative analysis. For decades classification (bottom row), entropy-based measures dominate: Entropy Ratio Order 1 leads at 17.04\%, Shannon Entropy follows at 16.49\%, and NRC Order 1 contributes 14.80\%. This entropy dominance, clearly visible in the tree-based importance plots, indicates that fine-grained temporal distinctions rely on information-theoretic complexity variations.

Centuries classification (top row of Figure~\ref{fig:compression_combined_analysis}) shows a different pattern. NRC Order 1 leads at 18.93\%, supported by NRC Order 2 at 16.92\%, while Shannon Entropy ranks third at 15.85\%. This preference for normalized compression measures, evident in both tree-based and permutation importance rankings, reflects their ability to capture broad historical transitions in linguistic regularity, structural changes that develop across extended periods. The correlation matrix (right column) reveals how different compression metrics relate to each other, with strong correlations between similar-order models while maintaining complementary discriminative power.

\subsection{Lexical Structure Features Domain Analysis}

Lexical and structural features capture stylistic conventions, syntactic complexity patterns, and formatting characteristics that vary across historical periods. Six features in this domain measure vocabulary diversity, sentence structure, word complexity, and surface-level formatting patterns reflecting both conscious stylistic choices and unconscious linguistic trends.

\begin{figure}[!h]
\centering
\includegraphics[width=1\textwidth]{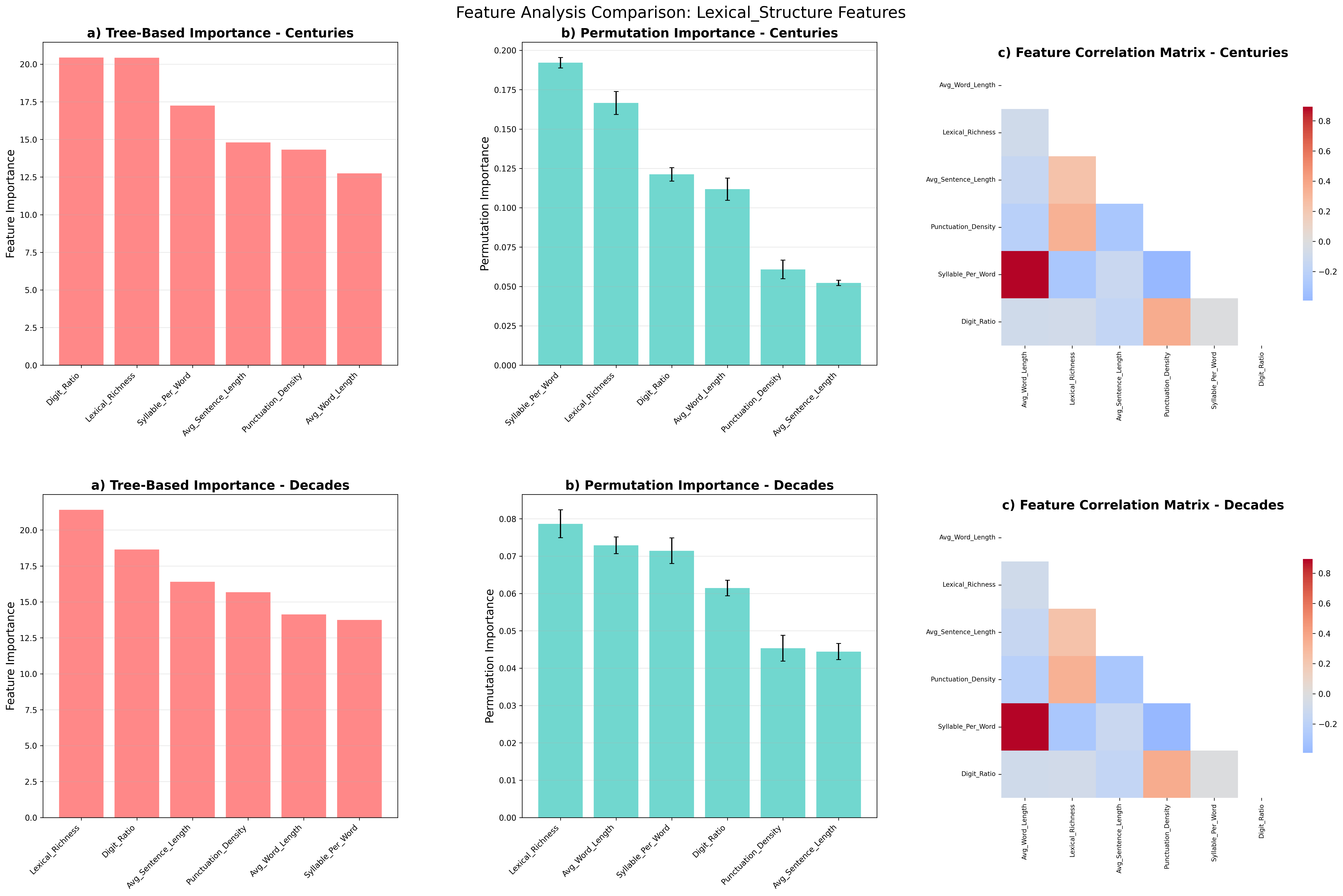}
\caption{Lexical structure features in centuries classification: a) Tree-based importance, b) Permutation importance, c) Feature correlations.}
\label{fig:lexical_structure_combined_analysis}
\end{figure}

As demonstrated in Figure~\ref{fig:lexical_structure_combined_analysis}, lexical structure features show consistent importance patterns across temporal scales. For centuries classification (top row), Digit Ratio (20.44\%) narrowly leads Lexical Richness (20.43\%) with only a 0.017\% margin, while decades analysis (bottom row) shows Lexical Richness at 21.40\%, placing Digit Ratio second at 18.64\%. This pattern, clearly visible in both importance ranking plots, establishes both vocabulary diversity and numerical content density as the most reliable temporal discriminators in this domain.

Average Sentence Length contributes significantly to temporal discrimination (16.40\% for decades, 14.80\% for centuries), reflecting documented historical trends in syntactic complexity. Educational standards, rhetorical conventions, and literacy levels influenced sentence construction patterns across centuries, creating measurable temporal signatures in structural complexity.

\FloatBarrier

\subsection{Distance Features Domain Analysis}

\begin{figure}[!h]
\centering
\includegraphics[width=1\textwidth]{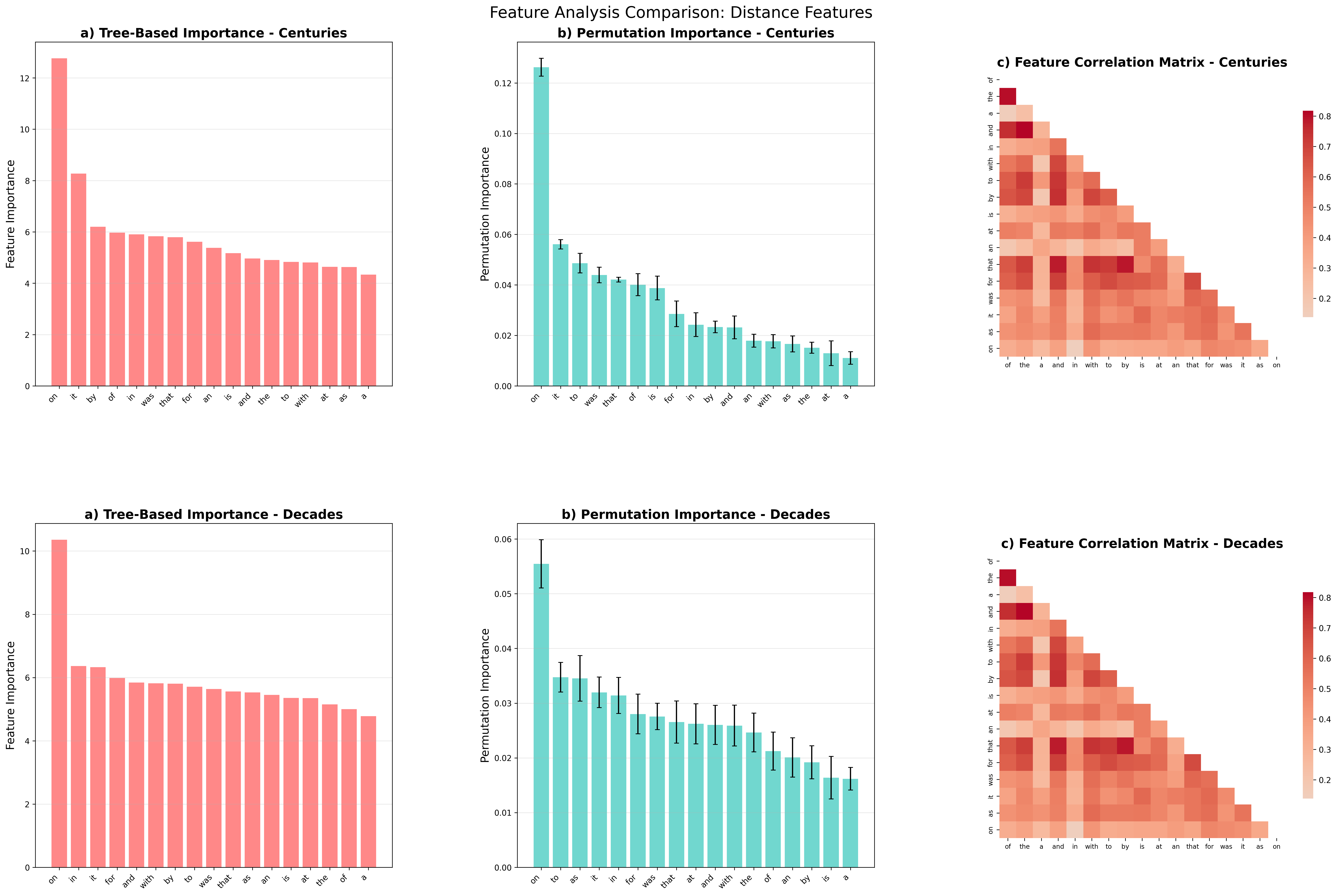}
\caption{Distance features in centuries classification: a) Tree-based importance, b) Permutation importance, c) Feature correlations.}
\label{fig:distance_combined_analysis}
\end{figure}

Distance features reveal grammatical change by tracking where function words appear in a sentence and how their spacing patterns shift as syntax evolves. We track positioning patterns for 17 function words representing core syntactic functions: prepositions, articles, auxiliary verbs, coordinating elements, and complementizers.
As illustrated in Figure~\ref{fig:distance_combined_analysis}, function word distance analysis reveals temporal patterns in syntactic organization. The feature importance plots clearly show that preposition distances dominate the rankings, with "on" providing the strongest temporal signal (10.36\% importance for decades classification), followed by "in" (6.37\%) and "of" (5.00\%). Article positioning patterns, especially "the" (5.15\%), capture changes in definiteness expression and noun phrase structure, while auxiliary verb distances reflect modifications in tense and aspect construction across historical periods. The correlation matrix reveals how different function words co-vary in their temporal patterns.

\FloatBarrier

\subsection{Neologism Detection Domain Analysis}

Neologism detection features capture temporal linguistic evolution through analysis of period-specific vocabulary and lexical innovation patterns. We use vocabulary development as one of the most distinct markers of historical linguistic change. Different periods bring terminology related to technological, social, cultural, and intellectual developments.

\begin{figure}[!h]
\centering
\includegraphics[width=1\textwidth]{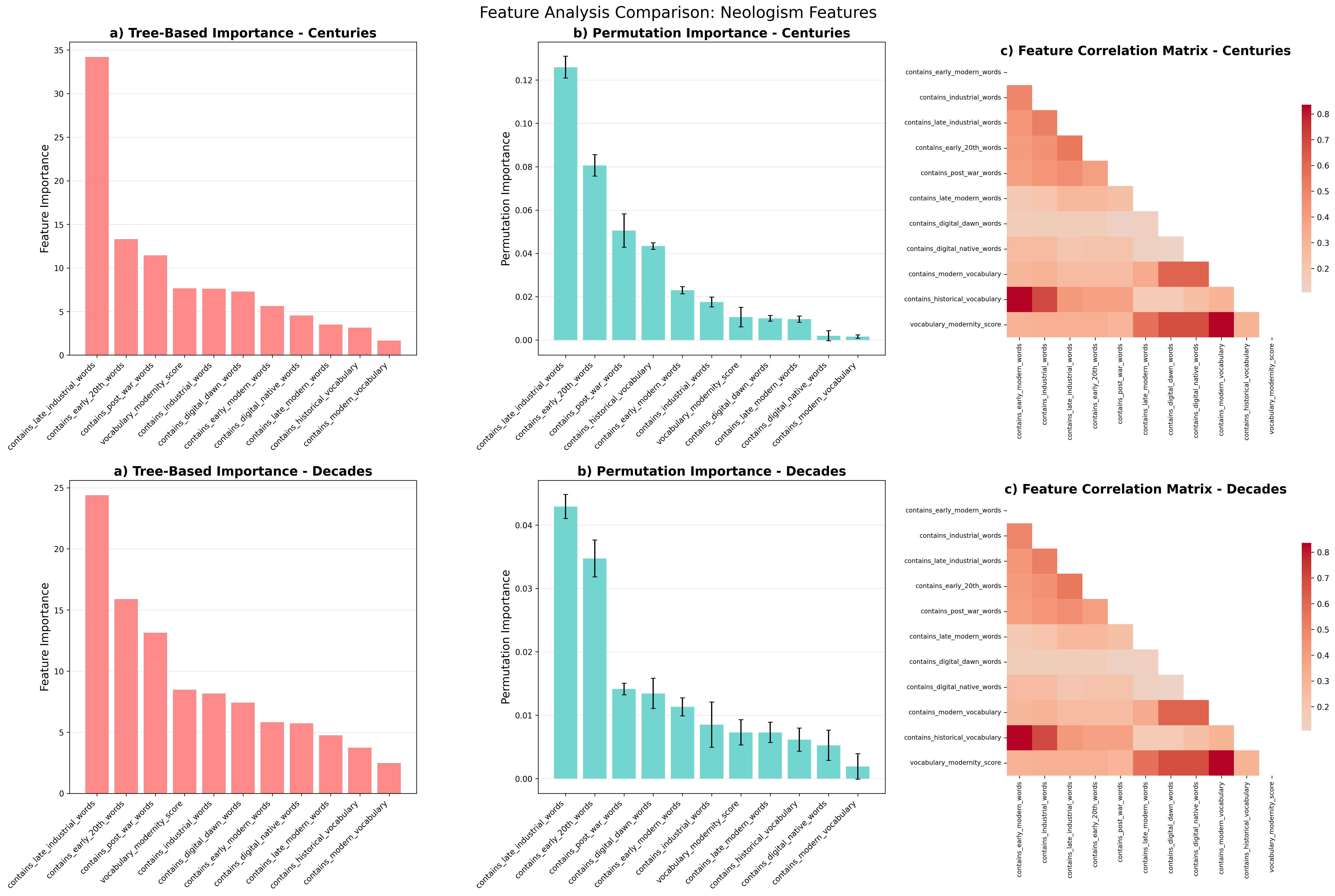}
\caption{Neologism features in centuries classification: a) Tree-based importance, b) Permutation importance, c) Feature correlations.}
\label{fig:neologism_combined_analysis}
\end{figure}

The neologism detection system tracks vocabulary evolution across eight historical periods, from the scientific revolution to the smartphone era. Eleven features capture how different periods introduced distinct terminology: boolean indicators for each historical period, aggregate features identifying broad temporal categories, and continuous measures quantifying vocabulary modernity proportions. As shown in Figure~\ref{fig:neologism_combined_analysis}, the late-industrial vocabulary feature dominates with 24.39\% importance for decades classification, clearly visible in the feature importance plots as representing approximately 2.5 times the contribution of the next strongest feature, highlighting the distinct vocabulary signatures of the Industrial Revolution period.

\FloatBarrier

\subsection{Readability Features Domain Analysis}

Readability features assess cognitive accessibility and complexity using established linguistic metrics that reflect historical trends in writing style, educational standards, and audience targeting. This domain employs two measures: Flesch Reading Ease scores and Stopword Ratio analysis.

Flesch Reading Ease provides difficulty measurement based on word and sentence length characteristics, capturing documented historical evolution in textual complexity. Stopword Ratio measures grammatical function word proportion, indicating syntactic density and stylistic complexity patterns that vary across historical periods.
These features show limited variation, with near 50\% importance distributed between them.

\subsection{Final Feature Analysis}

Cross-domain analysis reveals complementary strengths among feature categories that justify integrated modeling approaches. Compression features excel at capturing broad structural patterns and information-theoretic signatures. Lexical structure features provide reliable discrimination through vocabulary and sentence complexity. Function word distances reveal syntactic evolution patterns. Neologism detection captures explicit vocabulary innovation. Readability measures assess cognitive complexity evolution.

Performance evaluation shows integrated approaches significantly outperform individual domains, achieving complementary discrimination across multiple linguistic dimensions. This complementary effect is demonstrated across all comparative analyses in Figures~\ref{fig:compression_combined_analysis}, \ref{fig:lexical_structure_combined_analysis}, \ref{fig:distance_combined_analysis}, and \ref{fig:neologism_combined_analysis}, where different feature types capture distinct aspects of temporal linguistic evolution. 

Feature importance analysis across domains reveals temporal scale dependencies. Fine-grained decade classification benefits from entropy and complexity measures, while century-scale discrimination relies more on structural and vocabulary patterns. This scale-dependent optimization guides practical deployment for temporal classification applications requiring different granularity levels.

\subsection{SHAP Explainability Analysis}

SHAP analysis provides detailed insights into how individual features contribute to temporal predictions at both global model level and for individual texts. This analysis reveals that our models capture genuine linguistic evolution patterns rather than spurious correlations, with feature contributions aligning closely with established theories of historical language change.

The analysis demonstrates clear century-specific fingerprints across all feature domains, as illustrated in Supplementary Section~2. For compression features, the 17th century shows strong positive SHAP activations for NRC Order 1, reflecting the period's characteristic linguistic regularity and predictable structural patterns typical of early modern English with constrained vocabulary and formulaic constructions. The 18th century adopts a more distributed strategy, drawing balanced contributions from Shannon entropy and both NRC measures, suggesting a transitional period combining traditional linguistic regularity with emerging stylistic complexity. The 19th century exhibits strong activation for entropy ratio measures, capturing the industrial era's linguistic transformation through expanding technical vocabulary and evolving syntactic structures. Modern centuries (20th-21st) display inverted patterns with negative contributions from traditional compression measures alongside positive activation for complexity indicators, reflecting the emergence of varied, less predictable textual structures.

Lexical structure features reveal systematic temporal evolution patterns. Lexical Richness demonstrates the clearest progression, with SHAP values evolving from strongly negative contributions in the 17th century (reflecting constrained early modern vocabularies) to increasingly positive values in later periods, with particularly strong positive activation in the 19th century capturing the period's vocabulary explosion during industrialization. Average Word Length exhibits distinct signatures: the 18th century shows positive contributions for formal constructions, the 20th century demonstrates negative values reflecting modern trends toward direct expression, while the 21st century shows mixed patterns reflecting diverse digital communication styles. Syllables per Word reveals earlier centuries favoring syllabically complex formal language (positive SHAP contributions) transitioning to modern preferences for simplicity (negative contributions).

Distance features uncover sophisticated grammatical positioning evolution. The "on" feature demonstrates the most systematic temporal pattern, with strong positive SHAP contributions in the 17th century (indicating later sentence positions) gradually shifting to negative values in modern centuries, capturing fundamental grammatical reorganization toward direct subject-verb-object constructions. The 19th century shows particularly strong activation for auxiliary verb positioning ("is," "was"), capturing the industrial era's preference for complex predicate structures and formal register. Modern centuries exhibit inverted patterns where the "it" feature becomes important with positive SHAP values, reflecting increased use of pronoun-forward constructions in contemporary discourse.

Neologism features provide the most intuitively interpretable SHAP patterns. Late industrial vocabulary shows strong negative SHAP contributions in the 17th-18th centuries (reflecting logical absence of industrial terminology) and increasingly positive values in the 20th-21st centuries where such vocabulary becomes established. The 19th century exhibits unique characteristics as the vocabulary innovation epicenter, displaying balanced contributions across multiple neologism categories rather than extreme values, distinguishing texts from the coining period from those where industrial vocabulary had become fully integrated. Digital-era vocabulary categories show dramatic SHAP activation in the 21st century, with extreme positive values clearly separating contemporary texts from all historical periods.

Readability features demonstrate systematic evolution in writing complexity and accessibility. Flesch Reading Ease shows clear temporal progression with the 17th-18th centuries displaying strong positive SHAP contributions for higher complexity scores, reflecting these periods' preference for elaborate, formal constructions requiring advanced literacy. The 19th century emerges as a pivot point with balanced SHAP values capturing the coexistence of formal literary traditions alongside emerging popular writing styles. Modern centuries show negative SHAP contributions for complexity measures, indicating a shift toward more accessible, direct communication styles characteristic of contemporary writing.

The SHAP analysis validates several crucial aspects of our approach. First, it confirms that different feature domains capture complementary aspects of linguistic evolution, with each domain revealing distinct temporal signatures that align with historical linguistic theory. Second, it establishes the 19th century as a consistent pivot point across all feature domains, reflecting its role as a period of fundamental linguistic transition during industrialization. Third, it demonstrates that our models learn systematic rather than random patterns, with feature contributions following logical progressions that correspond to known processes of language change.

This interpretability transforms our tree-based models from black boxes into instruments for linguistic discovery, providing both accurate predictions and theoretical insights into the mechanisms of temporal linguistic variation.

%%%%%%%%%%%%%%%%%%%%%%%%%%%%%%%%%%%%%%%

\section{Results}
\label{sec:results}

Our experimental design evaluates temporal classification through multiclass prediction (identifying specific centuries or decades) and binary classification (distinguishing texts before/after a temporal threshold).

The dataset comprises texts from 1600-2020 across three corpora: validation and test sets from Open Library/Internet Archive, and Project Gutenberg for cross-domain evaluation. Six algorithms (CatBoost, XGBoost, Random Forest, SVM, k-NN, Gaussian Naive Bayes) are assessed using accuracy, F1-macro, AUC-ROC, AUPRC, MAE, and RMSE metrics.

Multiclass classification achieves 76.7\% accuracy for centuries and 26.1\% for decades, as detailed in Figure~\ref{fig:base_model_heatmap_accuracy}. Binary classification performance varies significantly based on the temporal boundary selection, with optimal performance around 1850-1900 thresholds as demonstrated in Figure~\ref{fig:binary_temporal_evolution}, indicating that linguistic change patterns are non-uniform across historical periods.

\subsection{Multiclass Classification Performance}

Predicting specific historical periods, rather than simply distinguishing old from new, presents greater challenges. These multiclass approaches aim for detailed temporal discrimination but face the difficulty of distinguishing between closely related time periods using linguistic cues alone.

\begin{figure}[!h]
\centering
\includegraphics[width=1.0\textwidth]{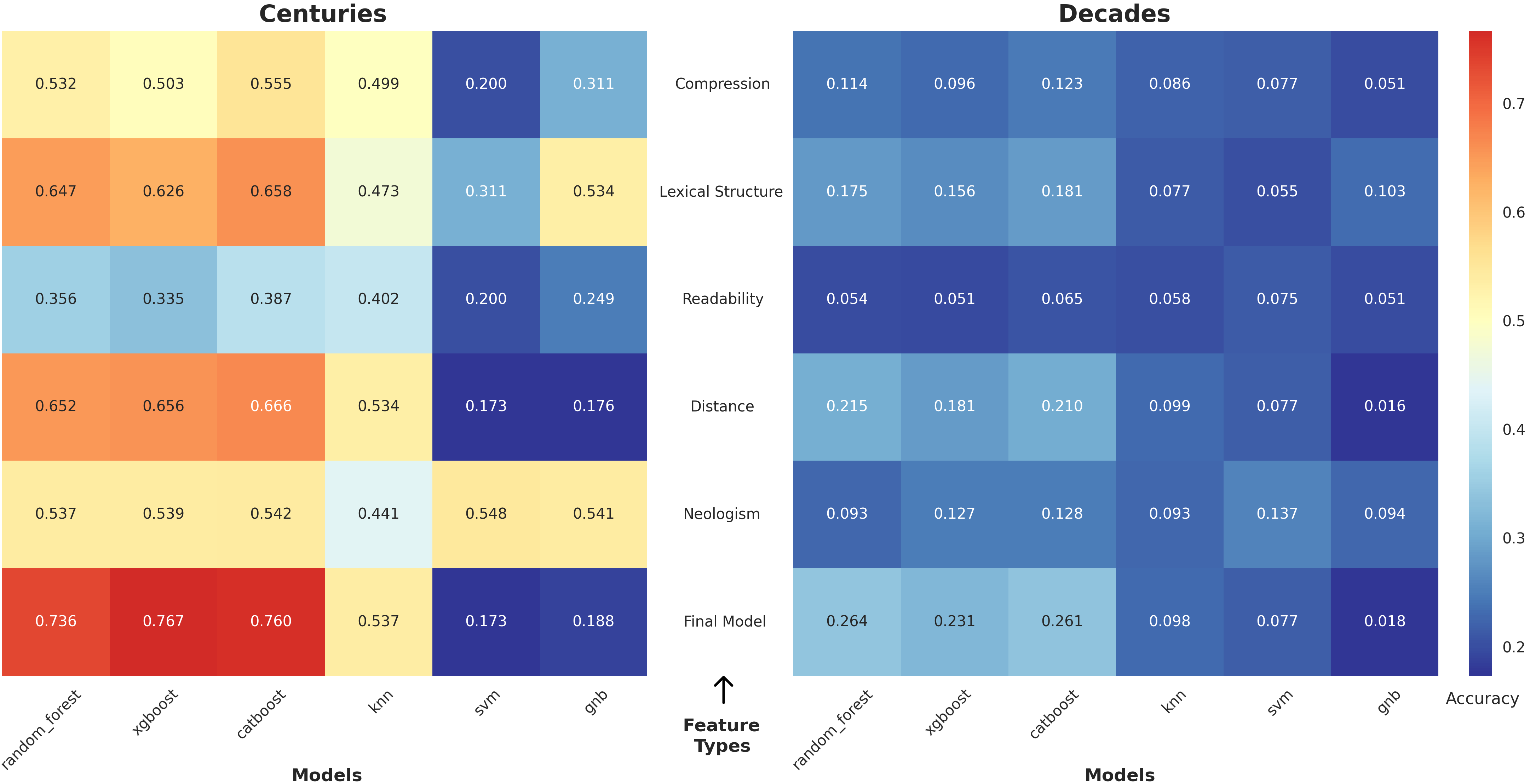}
\caption{Model accuracy for century (left heatmap) and decade (right heatmap) classification on the test dataset.}
\label{fig:base_model_heatmap_accuracy}
\end{figure}

Figure~\ref{fig:base_model_heatmap_accuracy} shows, for the test dataset, the classification accuracies of the final model and of models using individual feature types for predicting the century (left heatmap) and the decade (right heatmap).

Century-scale classification shows clear patterns in algorithmic and feature performance. The integrated final model yields our best result: XGBoost achieves 76.7\% accuracy on the test dataset, as shown in the rightmost column of Figure~\ref{fig:base_model_heatmap_accuracy}. This substantial improvement over any single feature domain, clearly visible when comparing the final model performance against individual feature columns, supports our hypothesis that combining multiple linguistic perspectives strengthens temporal classification.

Tree-based algorithms consistently outperform other approaches. CatBoost, XGBoost, and Random Forest achieve similar performance levels (73.6-76.7\% accuracy) with the full feature set. Individual feature domains show clear hierarchies. Distance features perform better than expected, achieving 65.2-66.6\% accuracy with tree-based models (visible in the "distance" column of Figure~\ref{fig:base_model_heatmap_accuracy}), suggesting that word spacing and proximity patterns carry temporal signals, consistent with the detailed analysis presented in Figure~\ref{fig:distance_combined_analysis}.

Lexical structure features perform nearly as well while showing greater algorithmic flexibility. Compression and neologism features occupy the middle tier, both reaching moderate effectiveness with tree-based models. Readability metrics underperform across both temporal scales, particularly at the decade level, suggesting that standard measures of text complexity may not capture the historical evolution of writing difficulty.

Decade-scale classification poses greater challenges. CatBoost achieves 26.1\% accuracy on the final model, an improvement over the 2.3\% random baseline but highlighting the difficulty of fine-grained temporal discrimination. The 43-class nature of this task, combined with class imbalance and limited per-class training data, creates computational and statistical challenges that even advanced algorithms struggle to overcome.

Traditional algorithms like SVM, KNN, and Gaussian Naive Bayes struggle across most feature types, with their best performance coming only with neologism features. At the decade scale, this algorithmic divide widens. Class imbalance amplifies the weaknesses of traditional approaches.

The ranking quality metrics reveal important insights about model reliability beyond simple accuracy scores.

\begin{table}[htbp]
\centering
\caption{Multiclass Performance Metrics: Century and Decade Classification (Test Dataset).}
\label{tab:comprehensive_multiclass_performance}
\begin{tabular}{lcccccc}
\toprule
\textbf{Feature Domain} & \textbf{Time Scale} & \textbf{Best Algorithm} & \textbf{Accuracy} & \textbf{F1-Macro} & \textbf{AUCROC} & \textbf{AUPRC} \\
\midrule
\multirow{2}{*}{\textbf{Distance}} & \textbf{Century} & \textbf{CatBoost} & \textbf{66.6\%} & \textbf{64.7\%} & \textbf{90.1\%} & \textbf{70.7\%} \\
                        & \textbf{Decade}  & \textbf{Random Forest} & \textbf{21.5\%} & \textbf{19.2\%} & \textbf{82.6\%} & \textbf{18.6\%} \\
\midrule
\multirow{2}{*}{\underline{Lexical Structure}} & \underline{Century} & \underline{CatBoost} & \underline{65.8\%} & \underline{64.6\%} & \underline{89.9\%} & \underline{68.3\%} \\
                                 & \underline{Decade}  & \underline{CatBoost} & \underline{18.1\%} & \underline{15.8\%} & \underline{85.2\%} & \underline{13.8\%} \\
\midrule
\multirow{2}{*}{Neologism} & Century & CatBoost & 54.2\% & 53.3\% & 82.5\% & 52.2\% \\
                         & Decade  & CatBoost & 12.8\% & 7.0\% & 75.2\% & 7.5\% \\
\midrule
\multirow{2}{*}{Compression} & Century & CatBoost & 55.5\% & 52.1\% & 82.8\% & 55.0\% \\
                           & Decade  & CatBoost & 12.3\% & 9.9\% & 75.8\% & 8.9\% \\
\midrule
\multirow{2}{*}{Readability} & Century & CatBoost & 38.7\% & 38.9\% & 72.2\% & 41.1\% \\
                           & Decade  & Random Forest & 5.4\% & 5.0\% & 61.5\% & 4.3\% \\
\midrule
\multirow{2}{*}{\texttt{Final Model}} & \texttt{Century} & \texttt{XGBoost} & \texttt{76.7\%} & \texttt{76.3\%} & \texttt{94.8\%} & \texttt{83.3\%} \\
                                    & \texttt{Decade}  & \texttt{CatBoost} & \texttt{26.1\%} & \texttt{23.6\%} & \texttt{90.4\%} & \texttt{23.6\%} \\
\bottomrule
\end{tabular}
\end{table}

AUCROC values consistently exceed 90\% for the final model across both temporal scales on test data, indicating strong discriminative capability for distinguishing between different historical periods. The final model achieves 94.8\% AUCROC for century classification and 90.4\% for decade classification, demonstrating that even when exact predictions fail, the models maintain excellent temporal ordering capabilities.

AUPRC metrics provide a more stringent assessment of performance, particularly sensitive to class imbalance effects inherent in historical text corpora. Century classification achieves 83.3\% AUPRC with the final model, while decade classification reaches 23.6\% AUPRC, a substantial gap that reflects the increased difficulty of maintaining precision-recall balance across 43 fine-grained temporal categories versus 5 broad centuries.

\subsection{Error Metrics Analysis}

Error metrics provide crucial insights into model reliability and failure patterns beyond simple accuracy scores. Table~\ref{tab:error_metrics_analysis} presents MAE and RMSE across feature domains, revealing distinct patterns in prediction reliability and failure modes.

For century-scale classification, the final model achieves the lowest error rates with MAE of 0.268 and RMSE of 0.592, representing an average absolute error of approximately 0.27 centuries. This substantial improvement over individual feature domains demonstrates the error-reduction benefits of multi-domain integration. Among individual features, lexical structure (MAE: 0.395, RMSE: 0.728) and distance features (MAE: 0.415, RMSE: 0.789) maintain the most controlled error patterns, while readability features exhibit the highest error variance (MAE: 0.959, RMSE: 1.359), indicating less reliable temporal discrimination.

The hierarchy of error performance closely mirrors accuracy rankings, with distance and lexical structure features providing the most reliable predictions when they fail. Compression features show moderate error rates (MAE: 0.631, RMSE: 1.063) with higher variance, suggesting occasional dramatic mispredictions that inflate RMSE relative to MAE. Readability features demonstrate poor error control across both metrics, with RMSE 32\% higher than MAE, indicating frequent large-magnitude errors.

\begin{table}[htbp]
\centering
\caption{Error analysis: MAE and RMSE performance across feature domains.}
\label{tab:error_metrics_analysis}
\begin{tabular}{lrrrr}
\toprule
\textbf{Feature Domain} & \multicolumn{2}{c}{\textbf{Century}} & \multicolumn{2}{c}{\textbf{Decade}} \\
\cmidrule(lr){2-3} \cmidrule(lr){4-5}
 & MAE & RMSE & MAE & RMSE \\
\midrule
Distance          & 0.415 & 0.789 & 4.67 & 7.99 \\
Lexical Structure & 0.395 & 0.728 & 4.18 & 6.66 \\
Neologism         & 0.534 & 0.853 & 5.17 & 7.69 \\
Compression       & 0.631 & 1.063 & 7.08 & 10.91 \\
Readability       & 0.959 & 1.359 & 10.65 & 14.09 \\
\midrule
\textbf{Final Model} & \textbf{0.268} & \textbf{0.592} & \textbf{3.03} & \textbf{5.33} \\
\bottomrule
\end{tabular}
\end{table}

Decade-scale classification reveals amplified error patterns reflecting the increased complexity of fine-grained temporal discrimination. The final model maintains relatively controlled errors with MAE of 3.03 decades and RMSE of 5.33 decades, representing reasonable temporal accuracy for applications tolerating moderate precision. The RMSE-to-MAE ratio of 1.76 indicates occasional severe mispredictions but generally controlled error variance.

Individual feature domains show dramatic error escalation at the decade scale. Readability features perform particularly poorly with MAE of 10.65 decades and RMSE of 14.09 decades. Compression features exhibit substantial errors with MAE of 7.08 decades and RMSE of 10.91 decades. The RMSE-to-MAE ratios reveal important variance patterns across feature domains. Readability features show the smallest ratio (1.32) despite their large absolute errors, while distance features exhibit the largest ratio among single-domain models (1.71), indicating more variable error behaviour. Lexical structure and compression features lie in between (1.59 and 1.54, respectively), combining relatively low absolute errors with moderate variance.

The error analysis confirms that temporal classification errors follow systematic rather than random patterns. The RMSE-to-MAE ratios across feature domains (ranging from 1.32 to 1.90) indicate that while typical errors remain moderate, occasional large-magnitude mispredictions contribute proportionally to overall error variance.
  
Distance and lexical structure features demonstrate the most reliable error characteristics with both low absolute errors and controlled variance, while readability and compression features exhibit higher error magnitudes and greater unpredictability. This pattern suggests that model confidence estimation could provide valuable deployment guidance for identifying predictions prone to large errors, particularly for compression and readability feature domains.

\subsection{Top-k Accuracy Analysis}

Being approximately right can be more valuable than being precisely wrong. Top-k accuracy analysis examines this trade-off by determining whether models that fail exact classification might still provide useful temporal guidance by ranking correct periods among their top predictions.

Century classification results are noteworthy: our final model improves from 76.7\% exact accuracy to 96.0\% top-2 accuracy, a 20 percentage-point improvement that indicates high reliability as shown in Table~\ref{tab:topk-decades-centuries}. When the model errs, the correct century almost always ranks as the runner-up, indicating temporal ordering rather than random confusion.

\begin{figure}[!h]
\centering
\includegraphics[width=1.0\textwidth]{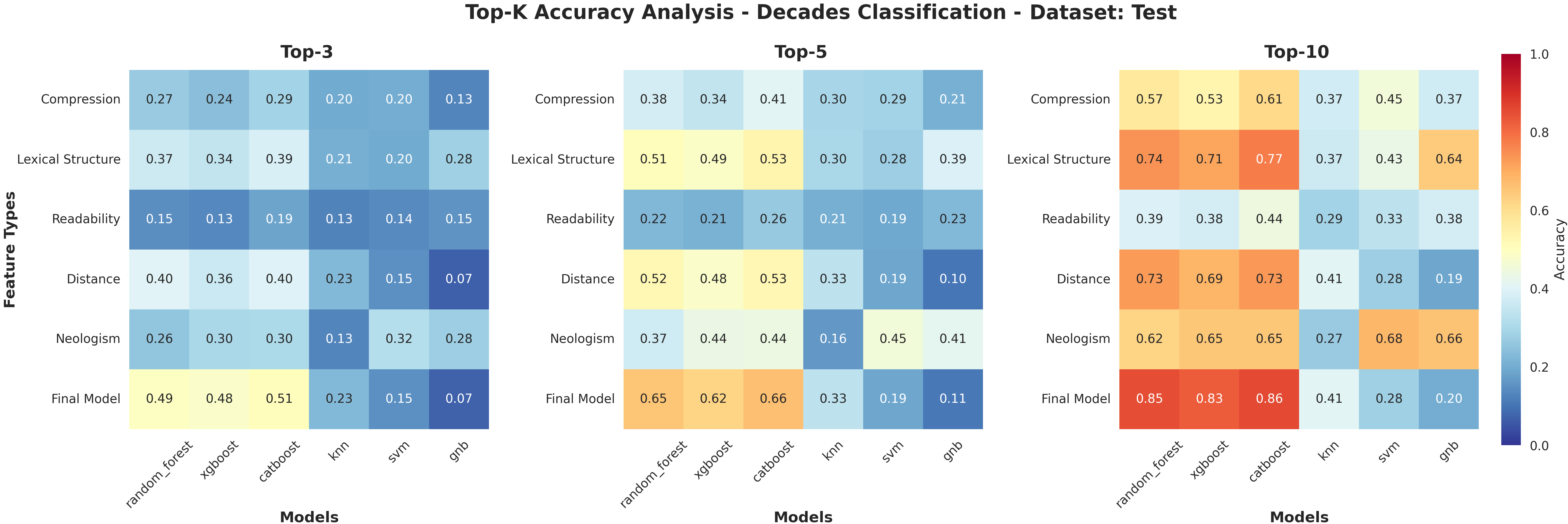}
\caption{Top-k accuracy analysis for decade classification showing performance improvement with relaxed temporal precision.}
\label{fig:topk_analysis}
\end{figure}

As depicted in Table~\ref{tab:topk-decades-centuries} and visualized in Figure~\ref{fig:topk_analysis}, decade classification is more challenging, with 43 temporal classes to distinguish. Analysis shows improvements through relaxed precision: our final model doubles its success rate in the top-3 predictions (50.8\% vs. 26.1\% exact), with top-5 accuracy reaching 66.4\% and top-10 accuracy reaching 85.8\%. These findings suggest practical deployment paths where accepting top-5 or top-10 predictions provides reliable temporal estimates with good confidence levels.

\begin{table}[htbp]
\centering
\caption{Top-k century and decade classification accuracy.}
\label{tab:topk-decades-centuries}
\begin{tabular}{lrrrr}
\toprule
\textbf{Metric} & \multicolumn{2}{c}{\textbf{Century}} & \multicolumn{2}{c}{\textbf{Decade}} \\
\cmidrule(lr){2-3} \cmidrule(lr){4-5}
 & Accuracy & Improvement & Accuracy & Improvement \\
\midrule
Top-1  & 76.7\% & ---      & 26.1\% & ---      \\
Top-2  & 96.0\% & +19.3\%  & ---    & ---      \\
Top-3  & ---    & ---      & 50.8\% & +24.7\%  \\
Top-5  & ---    & ---      & 66.4\% & +40.3\%  \\
Top-10 & ---    & ---      & 85.8\% & +59.7\%  \\
\bottomrule
\end{tabular}
\end{table}

\subsection{Binary Classification Results}

\begin{figure}[!h]
\centering
\includegraphics[width=1.0\textwidth]{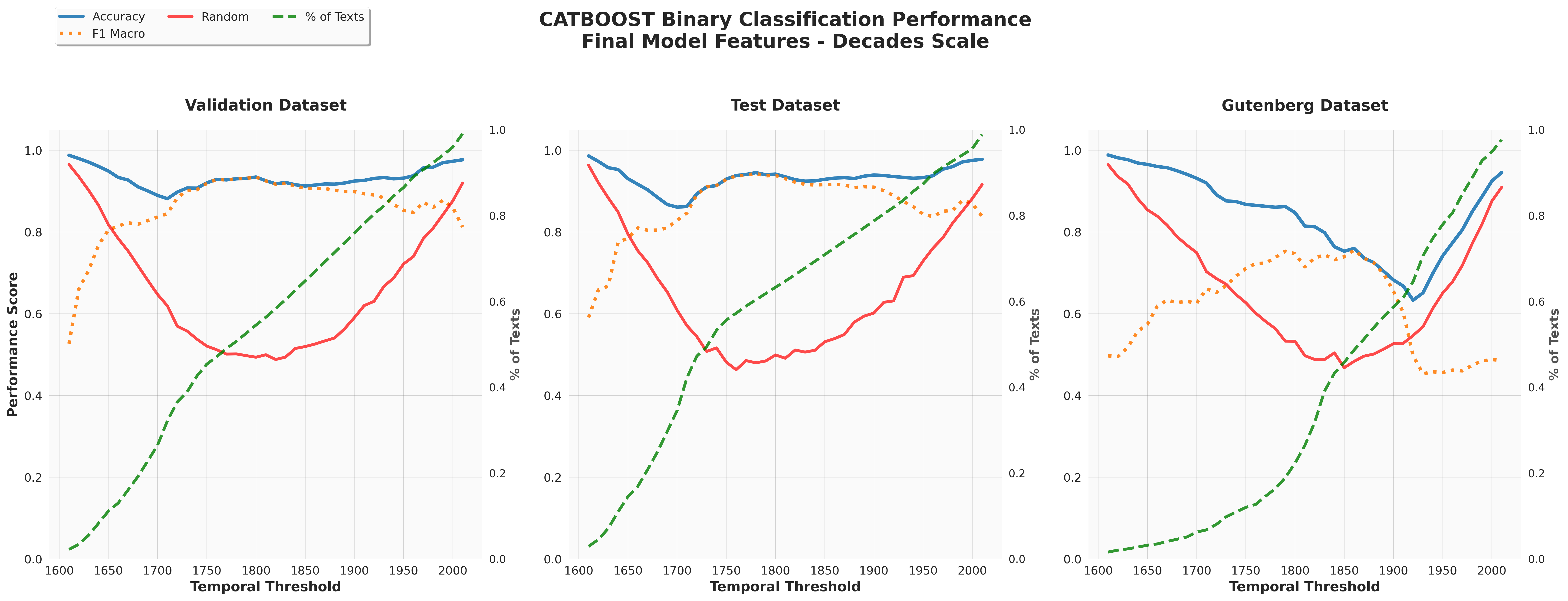}
\caption{Binary classification accuracy across decade thresholds.}
\label{fig:binary_temporal_evolution}
\end{figure}

Binary classification for temporal boundary detection achieved higher performance than multiclass classification, showing practical utility for applications requiring temporal threshold decisions. As shown in Figure~\ref{fig:binary_temporal_evolution}, the optimal temporal boundaries emerged around 1850-1900, achieving 85-98\% accuracy depending on the specific threshold and dataset.

The 1850-1900 period boundaries proved robust across datasets, with the 19th-20th century boundary emerging as a reliable choice for temporal discrimination. This boundary offers tolerance for boundary selection while maintaining good performance across different corpora, indicating generalization capabilities for historical transition detection. Class imbalance analysis revealed that boundaries in the 1850-1900 range provide better balance between historical and modern text proportions, avoiding the imbalances that affect earlier or later thresholds.

Cross-dataset evaluation shows that binary classification maintains better domain adaptation than multiclass approaches, with performance gaps of only 2-4 percentage points compared to 10-15 percentage points for decade-scale classification.

\subsection{Cross-Dataset Performance and Generalization}

Cross-dataset analysis reveals domain adaptation challenges across multiple dimensions of performance variation. While validation and test datasets align closely, the Gutenberg dataset's focus on literary texts creates biases in linguistic features. Literary language often employs archaic or formal registers that confound temporal boundary detection, explaining the difficulty in cross-domain generalization.

Performance gap analysis quantifies the underperformance of the Gutenberg dataset compared to test data: century classification drops from 76.7\% to 51.4\% accuracy (26.4 percentage point gap) with F1-macro gaps reaching 38.4 percentage points. Decade classification shows similar degradation from 26.1\% to 6.7\% accuracy (19.4 percentage point gap) with F1-macro gaps of 18.2 percentage points. These performance drops highlight the challenges of cross-domain temporal classification when moving from general historical texts to literary corpora.

\subsection{Algorithm Performance Patterns}

The most notable finding is the dominance of tree-based approaches. CatBoost, XGBoost, and Random Forest consistently outperform traditional methods, with advantages emerging for compression features and the integrated final model. Random Forest shows affinity for lexical structure features, while CatBoost excels with compression-based approaches.

Feature robustness varies across algorithmic families. Distance and neologism features prove relatively algorithm-agnostic, maintaining reasonable performance even with suboptimal methods. By contrast, compression features show sensitivity to algorithm choice, delivering meaningful results only with tree-based models. The AUPRC analysis confirms this pattern: robust feature domains sustain better precision-recall balance regardless of the chosen algorithm.

%%%%%%%%%%%%%%%%%%%%%%%%%%%%%%%%%%%%%%%%%%%%%%%%%%%

\section{Discussion}\label{sec:discussion}

Temporal text classification poses a challenging yet manageable computational problem for digital humanities and historical text analysis. Our findings reveal progress through feature engineering alongside inherent limitations in fine-grained temporal discrimination.

\subsection{Feature Domain Effectiveness and Linguistic Insights}

Feature performance reveals a clear hierarchy in temporal discrimination effectiveness. Distance features performed best among individual domains, achieving 66.6\% accuracy for century classification, with lexical structure features close behind at 65.8\%. This ranking indicates that syntactic evolution and vocabulary complexity provide the most reliable temporal signals in historical texts.

Feature combination produced marked improvements, validating the multi-faceted approach to temporal modeling. XGBoost achieved 76.7\% accuracy for century classification, over 10 percentage points better than any individual domain. Different linguistic aspects clearly contribute complementary signals that capture distinct dimensions of language evolution.

Compression-based features deserve attention despite their moderate performance (55.5\% accuracy). While insufficient alone for classification, they contributed meaningfully to the integrated model by providing theoretical grounding in information theory. These features capture temporal signatures related to text predictability and structural regularity that complement explicit linguistic measures.

\subsection{Temporal Resolution and Historical Complexity}

The stark contrast between century-scale (76.7\% with XGBoost) and decade-scale (26.1\% with CatBoost) classification accuracy reveals the inherent difficulty of fine-grained temporal discrimination. This gap likely reflects authentic linguistic phenomena rather than algorithmic limitations. Language evolves gradually, adjacent decades share many linguistic characteristics that make precise temporal boundaries difficult to establish.

The mid-19th century (1850-1900) proved challenging across multiple analyses, suggesting this period represents a linguistic transition zone. Rather than viewing these difficulties as failures, they may indicate periods where traditional and emerging linguistic forms coexisted, creating genuine ambiguity that mirrors human challenges in historical periodization.

Top-k accuracy analysis offers practical solutions for applications requiring temporal estimates rather than precise dating. The jump from 26.1\% exact accuracy to 85.8\% top-10 accuracy for decade classification shows models can provide reliable temporal guidance even when exact predictions fail, supporting real-world deployment.

\subsection{Cross-Dataset Challenges and Domain Adaptation}

Cross-dataset evaluation revealed domain adaptation challenges, with performance dropping when moving from general historical texts to literary corpora. These findings have implications for practical applications, temporal classification systems must account for genre and domain effects rather than assuming universal applicability.

\subsection{Algorithmic Performance and Interpretability}

Tree-based approaches dominated across all feature domains, validating their selection for temporal text classification. CatBoost, XGBoost, and Random Forest consistently outperformed traditional methods by wide margins, showing advantages for complex feature interactions and high-dimensional data typical of linguistic analysis.

Tree-based feature importance analysis and permutation importance evaluation provide interpretability, a crucial advantage for digital humanities applications. Understanding which linguistic features drive temporal predictions allows scholars to validate model decisions and gain insights into language evolution mechanisms. This transparency contrasts favorably with neural approaches that may achieve comparable accuracy but offer limited insight into underlying linguistic processes.

\subsection{Practical Implications for Digital Humanities}

The computational efficiency of our approach makes temporal analysis accessible to institutions with limited resources. Strong performance, interpretability, and modest computational requirements make this methodology suitable for widespread adoption in digital humanities contexts.

Binary classification results show practical utility for applications requiring temporal threshold decisions. Boundaries around 1850-1900 achieving 85-98\% accuracy provide reliable tools for distinguishing historical from modern texts, supporting authentication efforts and collection organization tasks.

The evaluation framework and feature importance analysis provide foundations for future research in computational temporal analysis. Identifying linguistic dimensions that drive temporal discrimination offers starting points for enhanced feature development and cross-linguistic investigation.

\subsection{Limitations and Methodological Considerations}

Several limitations warrant consideration for future applications and extensions. The focus on English texts limits generalizability to multilingual contexts, an important research direction for understanding universal versus language-specific temporal patterns. Historical text preservation biases affect learned patterns, with recent centuries better represented than distant periods.

Domain transfer challenges highlight the need for corpus-specific adaptation strategies. Future work should investigate domain adaptation techniques and develop methods for recognizing when temporal models encounter text types different from training distributions.

Feature robustness varies across algorithmic families, with some domains showing high sensitivity to algorithm choice while others maintain reasonable performance across methods. This suggests that feature selection strategies should consider not only discriminative power but also robustness to algorithmic variation and deployment constraints.

%%%%%%%%%%%%%%%%%%%%%%%%%%%%%%%

\section{Conclusion}\label{sec:conclusion}

This work demonstrates that historical text dating can be tackled effectively with interpretable, feature-engineered models. By combining compression-based, lexical, readability, neologism, and distance features with tree-based algorithms, we achieve 76.7\% accuracy at the century level and 26.1\% at the decade level, far above random baselines. When temporal precision is relaxed, performance further improves, with century-level top-2 accuracy reaching 96.0\% and decade-level top-10 accuracy reaching 85.8\%. Binary models around key historical thresholds (e.g., 1850–1900) attain 85–98\% accuracy, showing that rich temporal signals can be exploited in models that are both computationally efficient and easy to deploy.

Our analyses highlight clear strengths of the proposed framework. Distance and lexical structure features emerge as especially powerful, while compression-based metrics provide complementary information that improves overall performance. Explainability tools reveal temporal patterns in function word usage, vocabulary modernization, and readability that align with known historical transitions, illustrating how our models can serve as instruments for linguistic and historical insight rather than black boxes.

At the same time, the work exposes clear room for improvement. Cross-dataset evaluation with Project Gutenberg reveals domain-shift drops, and decade-level dating remains harder than century prediction. Strengthening the feature set, especially neologism models, and enlarging older corpora via probabilistic indexing \cite{toselli2024probabilistic,vidal2020carabela}, together with domain adaptation and multilingual extensions, offer tremendous potential.

%\section{Competing interests}
%The authors declare no competing interests.

\section*{Author contributions statement}

\textbf{P.P.:} Conceptualization, Methodology, Investigation, Software, Formal analysis, Writing, Review. \textbf{A.P.:} Conceptualization, Co-Supervision, Review. \textbf{D.P.:} Conceptualization, Methodology, Supervision, Writing, Review.

%\section*{Funding}
%Open access funded by Helsinki University Library. 

%\section{Acknowledgements}

%The authors wish to thank the Finnish Computing Competence Infrastructure (FCCI) for supporting this project with computational and data storage resources.

%\printbibliography
\bibliographystyle{abbrv}

\bibliography{references.bib}

%USE THE BELOW OPTIONS IN CASE YOU NEED AUTHOR YEAR FORMAT.
%\bibliographystyle{abbrvnat}
%\bibliography{reference}

\end{document}